# Towards the selection of patients requiring ICD implantation by automatic classification from Holter monitoring indices

Cappelaere CH[1,2], Dubois R[3], Roussel P[2], Baumann O[1], Amblard A[1], Dreyfus G[2]

[1]Sorin CRM SAS, Clamart, France
[2]SIGnal processing and MAchine learning (SIGMA) lab, ESPCI ParisTech, Paris, France
[3]IHU LIRYC, Bordeaux, France

## Abstract

*The purpose of this study is to optimize the selection of prophylactic cardioverter defibrillator implantation candidates. Currently, the main criterion for implantation is a low Left Ventricular Ejection Fraction (LVEF) whose specificity is relatively poor.*

*We designed two classifiers aimed to predict, from long term ECG recordings (Holter), whether a low-LVEF patient is likely or not to undergo ventricular arrhythmia in the next six months. One classifier is a single hidden layer neural network whose variables are the most relevant features extracted from Holter recordings, and the other classifier has a structure that capitalizes on the physiological decomposition of the arrhythmogenic factors into three disjoint groups: the myocardial substrate, the triggers and the autonomic nervous system (ANS). In this ad hoc network, the features were assigned to each group; one neural network classifier per group was designed and its complexity was optimized. The outputs of the classifiers were fed to a single neuron that provided the required probability estimate. The latter was thresholded for final discrimination*

*A dataset composed of 186 pre-implantation 30-mn Holter recordings of patients equipped with an implantable cardioverter defibrillator (ICD) in primary prevention was used in order to design and test this classifier. 44 out of 186 patients underwent at least one treated ventricular arrhythmia during the six-month follow-up period. Performances of the designed classifier were evaluated using a cross-test strategy that consists in splitting the database into several combinations of a training set and a test set. The average arrhythmia prediction performances of the ad-hoc classifier are NPV = 77% ± 13% and PPV = 31% ± 19% (Negative Predictive Value ± std, Positive Predictive Value ± std).*

*According to our study, improving prophylactic ICD-implantation candidate selection by automatic classification from ECG features may be possible, but the availability of a sizable dataset appears to be essential to decrease the number of False Negatives.*

## 1. Introduction

Sudden Cardiac Death (SCD) is an unexpected death caused by loss of heart function that occurs in a short time period (generally within one hour of symptom onset) in a person with known or unknown cardiac disease. Most SCDs are caused by a fast, erratic and disorganized propagation of impulses in the ventricles, named ventricular fibrillation. When it occurs, the heart is unable to pump blood anymore and death will occur within minutes, if left untreated by electrical shock(s).

Randomized clinical trials (MADIT II [1], SCD-HeFT [2]) have highlighted the benefits of prophylactic ICD implantations for SCD high-risk patients (post-Myocardial Infarction (MI) patients and Heart Failure (HF) patients with reduced LVEF).

However, according to [3], 81% of the patients have not received any therapy (appropriate or not) from their ICD over the 5-year follow-up period in SCD-HeFT. Beyond the economic issue caused by seemingly unnecessary implantations, are the health issues due to the peri- and postoperative complications. Thus, the selection of prophylactic ICD-implantation candidates must be improved.

In this study we propose the construction of a specific nonlinear classifier that relies on prior knowledge of the arrythmogenic factors, and uses the most relevant descriptors obtained from long-term ECG records (Holter) to identify patients who will undergo ventricular fibrillation in the next 6 months, hence are actually in need of a prophylactic ICD implantation.

## 2. Materials and Methods

### 2.1. Population study

One hundred and eighty-six patients (age 67±11 yrs, 163 males) with history of myocardial infarction and/or with heart failure and left ventricular dysfunctions (LVEF < 30%) have undergone a 30-mn Holter recording before being equipped with an ICD in primary prevention. During a six-month follow-up period, 44 out of 186 patients underwent at least one ventricular arrhythmia requiring a therapy deliverance from the ICD.

We divided the database into two groups: the positive group is composed of the 44 records that led to treated ventricular event in the next six months, and the negative group composed of the other 142 records.

## 2.2. Feature grouping and selection

Most of the known rhythmological and morphological parameters available from a Holter recording (such as the descriptors of the Heart Rate Variability, of the QT segment, of the QRS complex, etc.) are computed for each record of the database, resulting in a set of more than seventy candidate features.

These parameters describe different components which are implied in the arrhythmia genesis, so that they can be grouped by arrhythmogenic factors.

### 2.2.1 Feature grouping

The principal electrophysiological mechanism involved in the ventricular tachyarrhythmia genesis arises from the myocardial substrate, which refers to areas of fibrosis and ventricular dilatation. However, the substrate alone is not capable of originating tachyarrhythmia. The participation of trigger elements (the most common one is premature ventricular contraction (PVC)) is usually necessary. Additionally, the autonomic nervous system (ANS) interacts with the substrate and the triggers to cause electrical instability and leading to fatal arrhythmias, such as VF. Coumel schematized the interaction in the form of a triangle, each angle of which refers to one of the three factors (the myocardial substrate, the trigger elements and the autonomic nervous system) involved in the tachyarrhythmia genesis [4].

The structure of the second classifier that we have designed takes that prior physiological knowledge into account by processing separately the features pertaining to the three factors. Thus, the morphological features of the QRS complex, of the ST segment and of the T-wave, which describe the myocardial tissue state and the electrical conductivity condition, are grouped in the substrate hub; the occurrences of PVC and other rhythmic events are grouped in the triggers hub and the descriptors of the Heart Rate Variability and of the Heart Rate Turbulence, which characterize the autonomic regulation of the heart rate, are grouped in the ANS hub.

Nevertheless, due to the small amount of data compared to the large number of candidate features, any statistical model might be overly sensitive to noise or variance in the training data, and fail to estimate the underlying distribution from which the data were drawn. In other words, the model might overfit the training data. Overfitting usually leads to poor generalization capabilities of the classifier, i.e. to loss of accuracy on test (out-of-sample) data. In order to limit overfitting, a strategy of feature selection is proposed in the next section.

### 2.2.2 Feature selection

Within each hub, the most relevant features for a classification are selected by the random probe method [5]. This method ranks candidate features in order of decreasing relevance to predict ventricular arrhythmia, using Gram-Schmidt orthogonalisation. The originality of the random probe method lies in the addition of a pseudo-random variable (the probe) to the set of candidate features; its realizations are ranked just as all other candidate features. This results in an estimation of the risk $\rho$ of selecting a candidate feature although it might rank worse than an irrelevant variable, as a function of the number of selected candidate features.

## 2.3. Classifier design

We propose nonlinear classifiers that output an estimation of the probability for the patient to have a serious ventricular arrhythmia during the next six months.

These nonlinear classifiers are neural networks, all neurons of which have a sigmoid transfer function and the inputs of which are the features described in the previous section. In order to estimate the probability for a patient, given the inputs, to belong to the positive group [6], the samples were assigned the label 1 if the patients belonged to the positive group (i.e. had a treated ventricular arrhythmia), and 0 otherwise. Training was performed by gradient descent [7] followed by a BFGS [8] optimization of the least squares cost function with weight decay term [9]. The class imbalance problem was alleviated by multiplicating the records of arrhythmic patients in the training database.

The optimal complexity was found by $K$-fold cross-validation, whereby the training/validation set is split into $K$ homogenous and disjoint subsets, trainings are performed on $K$-1 subsets, and the mean squared error of the resulting models on the examples of the last ("validation") subset are computed; the procedure is iterated $K$ times, so that each example is in a validation subset once and only once. The cross-validation score is the average of the $K$ smallest validation mean squared errors. The complexity of the models that result in the smallest cross-validation score is selected.

After completion of complexity selection, the performances of the classifier and their variability are estimated as follows: the whole database is split into $K'$ homogenous and disjoint subsets, classifiers are trained on $K'$-1 subsets, the model that has the smallest training error is selected, and is applied to the data of the remaining ("estimation") subset; this procedure is iterated $K'$ times in order to use each example once and only once in the estimation set. The $K'$ classification performances are averaged and their standard deviation is computed.

### 2.3.1. Conventional neural network classifier

As a reference, a single hidden layer neural network classifier was designed, the inputs of which were the eighteen features selected as described in section 2.2.2 (Figure 1).

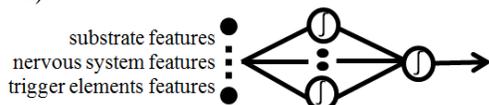

Figure 1. Single hidden layer neural network classifier.

### 2.3.2. Ad hoc neural network classifier

In order to capitalize on prior knowledge, an ad hoc network was designed by grouping the variables as described in the section 2.2.1. Each group of factors undergoes a distinct nonlinear transformation, whose results are fed to a non-linear neuron that provides an estimate of the risk pertaining to the patient (Figure 2).

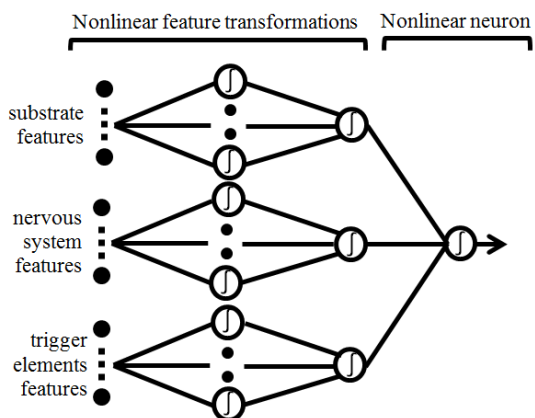

Figure 2. Classifier composed of three subnetworks linked to an output nonlinear neuron.

The first step was to build one neural network per arrhythmogenic factor, with one hidden layer and one output neuron. The optimal complexity for each of these three subnetworks is found by cross-validation.

Finally, the output of each subnetwork is fed to a non-linear neuron. The overall network is trained, the parameters of the subnetworks, obtained in the previous step, are taken as initial values, and the classifier that has the best performance on the training/validation set is selected.

## 3. Results

The most relevant features (with a risk $\rho \leq 10\%$) for discrimination between patients likely or not to undergo ventricular tachyarrhythmia in the next 6 months are listed, per hub, in Table 1.

Table 1. Most relevant features for classification.

| | |
|---|---|
| Myocardial Substrate | QRS residuum † |
| | QRS-T angle † |
| | T-Wave residuum † |
| | QT end † |
| | QT apex † |
| | QT slope † |
| | ST elevation / depression † |
| Autonomic Nervous System | Minimum heart rate † |
| | Mean RR interval † |
| | SDANN |
| | Poincaré Plot Analysis SD2 |
| | Heart rate variability index ‡ |
| | Turbulence Onset |
| Trigger elements | Ventricular bigeminy |
| | Ventricular trigeminy |
| | Non-sustained ventricular tachycardia |
| | Premature Atrial Contraction (PAC) |
| | Couple of PAC |

† averaged over the signal duration
‡ heart rate variability index is the percentage of averaged differences between two successive complexes.

These features were available for 107 patients out of 186.

The selected structure for the conventional single hidden layer neural network had one hidden neuron in its hidden layer.

Concerning the ad hoc network, with three hubs of inputs, it was found that the most appropriate nonlinear transformation, for each subnetwork, was performed by a single neuron with sigmoid output. In other words, the subnetworks shown on Figure 1 had no hidden neuron, i.e. performed a linear separation between the classes.

The performance of the conventional classifier, estimated as described in section 2.3 with $K'$ = 10 resulted in an average reduction of the ICD-implantation of 52% (std. 19%). The averaged negative predictive value (NPV) assessed on cross-test was 68% (std. 13%); it is the ratio of the number of patients who are correctly classified as not requiring an ICD-implantation to the number of patients who were classified as not requiring it. The averaged positive predictive value (PPV), which is the proportion of patients rightly classified as needing an ICD-implantation among the patients classified as requiring it., was 25% (std. 20%).

Likewise, the estimated performances obtained by the ad hoc classifier were an averaged reduction of the ICD-

implantations of 59% (std. 15%), with an average NPV of 77% (std. 13%) and a PPV of 31% (std. 19%).

In other words, let us build, for each classifier, an overall confusion matrix by summing the ten matrices produced in the performance estimation procedure. With this "theoretical" classifier, based on the performances of the conventional classifier, only 52 out of the 107 patients would be implanted, resulting in a 51.4% reduction of the number of implantations. Among patients for whom implantation is recommended by the classifier, 13 actually require it, resulting in a PPV of 25%, and among the 55 patients for whom implantation is rejected, 39 do not require it, resulting in a NPV of 71% (Table 2).

In the same situation, a "theoretical" classifier based on the performances of the ad hoc classifier would recommend a reduction of ICD implantations by 58.9% Among the 44 patients for whom an implantation is recommended, 14 really need it, thus the PPV is equal to 32%. Among the 63 patients for whom the implantation is considered as unrequired, 48 do actually not need it, which reflects a NPV of 76% (Table 3).

Table 2. Performances of the conventional classifier.

| Overall confusion matrix | | "Theoretical" classifier | Designed classifiers |
|---|---|---|---|
| 39 | 39 | NPV = 71% | NPV = 68±13% |
| 16 | 13 | PPV = 25% | PPV = 25±20% |
| Correctly classified patients | | | 5.1 ± 2.0 |

Table 3. Performances of the ad hoc classifier.

| Overall confusion matrix | | "Theoretical" classifier | Designed classifiers |
|---|---|---|---|
| 48 | 30 | NPV = 76% | NPV = 77±13% |
| 15 | 14 | PPV = 32% | PPV = 31±19% |
| Correctly classified patients | | | 6.2 ± 1.5 * |

* conventional vs ad hoc classifiers: p-value = 0.0547

## 4. Discussion

In this study, the objective was to reduce the rate of ICD-implantation with more than 90% of negative predictive value and at least 20% of positive predictive value. The reduction of ICD implantations made possible by both of the designed classifiers is noteworthy and the desired PPV is obtained in both cases. Furthermore, in this experimental case, the contribution of the decomposition of the inputs according to their arrhythmogenic participation provides a slight improvement (p-value = 0.0547) but is not statistically significant. Nonetheless, the aim in NPV remains out of reach considering the limitations we have to face.

The database contained few examples and was unbalanced, making the learning of the training set features difficult, regardless the complexity of the network.

Another limitation of the database is the length of the recordings, which was only 30 minutes, at any time in the daytime; it was thus impossible to study some time periods that would be of interest, such as the hour before awakening. Furthermore, it has been impossible to calculate some descriptors in the usual way (for example, some descriptors of the HRV are commonly averaged on 24 hours) and a temporal analysis of the descriptor variations was unfeasible.

Therefore, repeating the same process of classifier construction on a sizable database of 24-hr pre-implantation Holter recordings of patients equipped with an ICD in primary prevention seems to be mandatory.

## 5. Conclusion

Improving prophylactic ICD-implantation candidate selection by automatic classification from ECG features may be possible. Nevertheless, to reach this aim, getting more suitable and larger databases is essential to decrease the number of False Negatives, hence increase the negative predictive value.

Address for correspondence.
Charles-Henri Cappelaere, Sorin CRM
4 av. Réaumur, 92140 Clamart, France
ch.cappelaere@gmail.com